\documentclass[conference]{IEEEtran}
\usepackage{graphicx,times,amsmath,amssymb,cite,subfigure,stfloats,booktabs,multirow}
\usepackage[lined,ruled,commentsnumbered]{algorithm2e}

\makeatletter

\newcommand{\Rmnum}[1]{\expandafter\@slowromancap\romannumeral #1@}
\makeatother

\begin{document}

\title{Active Learning for Black-Box Adversarial Attacks in EEG-Based Brain-Computer Interfaces}

\author{\IEEEauthorblockN{Xue Jiang, Xiao Zhang and Dongrui Wu}
\IEEEauthorblockA{School of Artificial Intelligence and Automation\\
Huazhong University of Science and Technology, Wuhan, China\\
Email: xuejiang@hust.edu.cn, xiao\_zhang@hust.edu.cn, drwu@hust.edu.cn}}

% conference papers do not typically use \thanks and this command
% is locked out in conference mode. If really needed, such as for
% the acknowledgment of grants, issue a \IEEEoverridecommandlockouts
% after \documentclass

% use for special paper notices
%\IEEEspecialpapernotice{(Invited Paper)}

% make the title area
\maketitle

% in the abstract
\begin{abstract}
Deep learning has made significant breakthroughs in many fields, including electroencephalogram (EEG) based brain-computer interfaces (BCIs). However, deep learning models are vulnerable to adversarial attacks, in which deliberately designed small perturbations are added to the benign input samples to fool the deep learning model and degrade its performance. This paper considers transferability-based black-box attacks, where the attacker trains a substitute model to approximate the target model, and then generates adversarial examples from the substitute model to attack the target model. Learning a good substitute model is critical to the success of these attacks, but it requires a large number of queries to the target model. We propose a novel framework which uses query synthesis based active learning to improve the query efficiency in training the substitute model. Experiments on three convolutional neural network (CNN) classifiers and three EEG datasets demonstrated that our method can improve the attack success rate with the same number of queries, or, in other words, our method requires fewer queries to achieve a desired attack performance. To our knowledge, this is the first work that integrates active learning and adversarial attacks for EEG-based BCIs.
\end{abstract}


\section{Introduction} \label{sect:Intro}
% EEG-based BCI
A brain-computer interface (BCI) is a communication system that connects a human brain and a computer~\cite{BCIIntro}. Direct dialogue between the brain and the computer can be achieved by the established information path. Electroencephalogram (EEG) is the most frequently used input signal in BCIs, due to its low cost and convenience~\cite{BCIReview}. Various paradigms are used in EEG-based BCIs, such as P300 evoked potentials~\cite{P300Sutton, P3001988,drwuTHMS2017,drwuTNSRE2016}, motor imagery (MI)~\cite{MI2001}, steady-state visual evoked potential (SSVEP)~\cite{SSVEPSurvey}, etc.

% Deep learning and adversarial examples
Deep learning has achieved great success in numerous fields. Multiple convolutional neural network (CNN) classifiers have also been proposed for EEG-based BCIs. Lawhern \emph{et al.}~\cite{EEGNet} proposed EEGNet, which can be applied to different BCI paradigms. Schirrmeister \emph{et al.}~\cite{MNE} designed a deep CNN model (DeepCNN) and a shallow CNN model (ShallowCNN). In addition, there were some studies to convert EEG signals into images and then classify them with deep learning models~\cite{EEG2Image, DeepLearningMI, Tayeb2019}. This paper considers only CNN models (i.e., EEGNet, DeepCNN, ShallowCNN) which take the raw EEG signals as the input.

Despite their state-of-the-art performance, recent studies have shown that deep learning models are vulnerable to adversarial examples, which are crafted by adding small imperceptible perturbations to benign examples to degrade the performance of a well-trained deep learning model. For example, in face recognition, an adversarial perturbation can be attached to the glasses, and the attacker who wears it can avoid being recognized, or be recognized as another person~\cite{FaceAttack}. In image classification, the adversarial examples can fool a deep learning model to give incorrect image labels~\cite{FGSM, AdvExamSzegedy, BIM, Adv3D}. Many adversarial attacks have also been performed in speech recognition~\cite{AdvExamAudio}, malware classification~\cite{Malware2016}, semantic segmentation~\cite{Semantic2017}, etc. Recently, Zhang and Wu~\cite{zhang2019vulnerability} verified that adversarial examples exist in EEG-based BCIs and proposed several adversarial attack approaches.

% Black-box attack and transferability-based method
Many effective algorithms for generating adversarial examples have been proposed, such as the fast gradient sign method (FGSM)~\cite{FGSM}, the C\&W method~\cite{AdvCW}, L-BFGS~\cite{AdvExamSzegedy}, the basic iterative method~\cite{BIM}, DeepFool~\cite{Moosavi-Dezfooli2016}, etc. These methods mainly considered the white-box attack scenario, where the attacker has full access to the target model, including its architecture and parameters. Accordingly, the attacker can perform attacks by adding perturbations along the direction calculated by gradient-based strategies or optimization-based strategies. However, the white-box setting requires full information of the target model, making it impractical in many real-world applications.

In this paper, we focus on a more realistic and challenging black-box attack scenario, where the attacker can only observe the target model's responses to inputs but has no information about its architecture, parameters, and training data. The attacker needs to generateadversarial examples whose perturbations are constrained to a magnitude threshold with limited queries. Papernot \emph{et al.}~\cite{BlackBoxAttack} proposed a black-box attack method which generated adversarial examples for a white-box substitute model and attacked the black-box target model based on the transferability. Zhang and Wu~\cite{zhang2019vulnerability} proposed an unsupervised fast gradient sign method (UFGSM) to craft adversarial examples for black-box attacks in EEG-based BCIs. However, these transferability-based approaches suffered from low query efficiency: they usually require a large number of queries to build a substitute model that is sufficiently similar to the target model.

% Active Learning
To address this issue, we introduce a query synthesis based active learning strategy to transferability-based black-box attacks of EEG-based BCIs. Given a small amount of initial training EEG epochs for the substitute model, we first randomly obtain a pair of opposite instances in each iteration, i.e., two instances of different classes. According to the initial opposite-pair, we use a binary search strategy to synthesize another opposite-pair close to the current classification boundary in the input space. After that, we synthesize queries along the perpendicular line of the previously found opposite-pair. This query synthesis based active learning strategy can directly synthesize queries which are close to the target model's decision boundary and well scattered. It improves the query efficiency by directly searching for informative examples in the input space for substitute model training, instead of taking a fixed step along the gradient direction. Experiments on three CNN classifiers and three BCI datasets demonstrated the effectiveness of our proposed method.

The remainder of this paper is organized as follows: Section~\ref{sect:RelatedWork} introduces related work on black-box adversarial attacks and active learning. Section~\ref{sect:Methodology} proposes our query synthesis based active learning approach for crafting adversarial examples for black-box attacks in EEG-based BCIs. Section~\ref{sect:Experiment} evaluates the attack performance of our proposed approach. Finally, Section~V draws conclusion.

\section{Related Work} \label{sect:RelatedWork}

In this section, we briefly review previous studies on black-box attacks and active learning.

\subsection{Black-Box Attacks}

Black-box attacks can be roughly divided into three categories: decision-based, score-based and transferability-based. Decision-based attacks were first proposed by Brendel \emph{et al.}~\cite{DecisionBasedAttack}. Its main idea is to gradually reduce the magnitude of the adversarial perturbation while ensuring its effectiveness. Score-based attacks rely on the model's output scores, e.g., class probabilities or logits, to estimate the gradients and then generate adversarial examples~\cite{ilyas2018black, ScoreBased2018}. Transferability-based attacks were first proposed by Papernot \emph{et al.}~\cite{BlackBoxAttack} in image classification. The attacker trains a substitute model, which solves the same classification problem as the target model, to generate adversarial examples for the target model.

In transferability-based black-box attacks, the key step is to learn a substitute model whose decision boundary resembles the target model's. Papernot \emph{et al.}~\cite{BlackBoxAttack} used Jacobian-based dataset augmentation to synthesize a substitute training set and labeled it by querying the target model. By alternatively augmenting the training set and updating the substitute model, it can gradually approximate the target model. Recently, Zhang and Wu~\cite{zhang2019vulnerability} extended this idea to EEG-based BCIs, but their approach was slightly different: they synthesized a new training set by using the loss computed from the inputs, instead of the labels from the target model, to calculate the Jacobian matrix.

Albeit the outstanding attack performance of these methods, they always require a large number of queries to train a substitute model. Generally, the number of queries grows exponentially with the number of iterations. In order to improve the query efficiency, we propose an active learning based data augmentation approach to train the substitute model in transferability-based black-box attacks.

\subsection{Active Learning}

Active learning is an effective way to reduce the data labeling effort, by actively selecting the most useful instances to label. There are two main scenarios of active learning in the literature: query synthesis~\cite{query1988queries, query2009automation, query2015efficient, query2015active} and sampling. The latter can be further divided into stream-based sampling~\cite{stream1989training, stream2008general} and pool-based sampling~\cite{sample2008analysis, wu2016offline, wu2019active, wu2019affect}. Sampling-based active learning selects real unlabeled instances from a pool or steam for labeling. In query synthesis based active learning, one can query any data instance in the input space, including synthesized instances.

Intuitively, query synthesis can be applied to the training process of the substitute model to improve the query efficiency of black-box attacks, because it can actively synthesize more informative EEG epochs than generating some epochs in Jacobian-based way. Furthermore, compared with sampling, it is more efficient to synthesize a query directly instead of evaluating every instance in an unlabeled data pool. Here we don't need to worry about the fact that some synthesized EEG epochs may be unrecognizable to human~\cite{query1992query}, because the target model can label any instance in the input space. Our ultimate goal is to identify the decision boundary of the target model by a minimum number of queries.

\section{Methodology} \label{sect:Methodology}

In this section, we first introduce the transferability-based black-box attack setting for CNN classifiers in EEG-based BCIs, and then describe the query-synthesis-based active learning strategy for training the substitute model, and adversarial example crafting for the target model.

\subsection{Attack Setting}

The attack framework in this paper is the same as our previous work~\cite{zhang2019vulnerability}, where the attackers can add adversarial perturbations before the machine learning modules.

Let $\mathbf{x}_i\in \mathbb{R}^{C \times T}$ be the $i$-th raw EEG epoch ($i=1,...,n$), where $C$ is the number of EEG channels and $T$ the number of the time domain samples. Let $f_{\theta}(\mathbf{x}_i)\rightarrow y_i$ denote the CNN model that predicts the label for an input EEG epoch.

Given a target CNN model and a normal EEG epoch $\mathbf{x}_i$, the task is to generate an adversarial example $\mathbf{x}_i^*$ misclassified by the CNN model. Formally, an adversarial example $\mathbf{x}_i^*$ should satisfy the following constraints:
\begin{align}
&f_{\theta}(\mathbf{x}_i^*) \neq y_i, \label{eq:limitation}\\
&D(\mathbf{x}_i, \mathbf{x}_i^*) \leqslant \epsilon, \label{eq:bound}
\end{align}
where $D(\cdot,\cdot)$ is a distance metric, and $\epsilon$ constrains the magnitude of the adversarial perturbation. (\ref{eq:limitation}) ensures the success of the attack, and (\ref{eq:bound}) ensures that the perturbation is not larger than a predefined upper bound $\epsilon$.

Next we describe an algorithm to learn a substitute model for a given target CNN classifier by querying it for labels on query-synthesized inputs, and then introduce the UFGSM approach~\cite{zhang2019vulnerability} to craft adversarial examples on the trained substitute model, which can also be transferred to the target model. We start with binary classification, and then extend it to multi-class tasks.

\subsection{Query-synthesis-based Dataset Augmentation for Training the Substitute Model}

In the black-box attack scenario, we have no access to the architecture, parameters and training data of the target model, but can input EEG trials to the target model and observe the corresponding outputs to probe the model. As in~\cite{BlackBoxAttack}, we also adopt the transferability-based approach to implement the black-box attack. The difference is that we use a query-synthesis-based active learning strategy as the data augmentation technique in substitute model training rather than the Jacobian-based approach. The oracle in query synthesis active learning is the target model.

\subsubsection{Binary search synthesis}

Assume a small labeled training set $S_0$ has been obtained from querying the target model. An initial substitute model $f_0'$ can be trained on this set. Suppose $\{\mathbf{x}_0^+,\mathbf{x}_0^-\}$ is an opposite-pair in $S_0$. Then, we query their middle point on the substitute model to find another opposite-pair closer to the decision boundary, using Algorithm~\ref{alg:BS}.

\begin{algorithm}
\DontPrintSemicolon
\KwIn{$\{\mathbf{x}_0^+,\mathbf{x}_0^-\}$, initial opposite-pair of EEG epochs; $f'$, current substitute model; $m$, maximum number of binary search iterations.}
\KwOut{$\{\mathbf{x}^+,\mathbf{x}^-\}$, an opposite-pair of EEG epochs.\\
~\\}
$\mathbf{x}^+=\mathbf{x}_0^+$;\\
$\mathbf{x}^-=\mathbf{x}_0^-$;\\
\For{$i=1$ to $m$}
{
    $\mathbf{x}_b=(\mathbf{x}^{+}+\mathbf{x}^-)/2$;\\
    Query $f'$ for $y_b$, the label of $\mathbf{x}_b$;\\
    \uIf{$y_b$ is positive}{
        $\mathbf{x}^+ \leftarrow \mathbf{x}_b$ \;}
    \Else{
        $\mathbf{x}^- \leftarrow \mathbf{x}_b$ \;}
}
~\\
\KwRet $\{\mathbf{x}^+,\mathbf{x}^-\}$
\caption{
    Binary search synthesis.
    $\{\mathbf{x}^+,\mathbf{x}^-\}=BinarySearch(\{\mathbf{x}_0^+,\mathbf{x}_0^-\},f',m)$
}\label{alg:BS}
\end{algorithm}

\subsubsection{Mid-perpendicular synthesis}

There is an obvious limitation in binary search synthesis: if we always use binary search to generate training epochs, they may concentrate in one area and lack diversity. Therefore, we synthesize the next query along the mid-perpendicular direction after we find an opposite-pair close enough to the decision boundary. Specifically, we find the opposite-pair's orthogonal vector by Gram-Schmidt process~\cite{query2015active}, set the magnitude of the orthogonal vector to $q$, then move it to a more precise midpoint. The details are shown in Algorithm~\ref{alg:Perp}.

\begin{algorithm}
\DontPrintSemicolon
\KwIn{$\{\mathbf{x}_b^+,\mathbf{x}_b^-\}$, an opposite-pair of EEG epochs; $f'$, current substitute model; $m$, maximum number of binary search iterations; $q$, magnitude of the orthogonal vector.}
\KwOut{$\mathbf{x}_s$, an synthesized EEG epoch.\\
~\\}
$\mathbf{x}_1=\mathbf{x}_b^{+}-\mathbf{x}_b^-$;\\
Generate an EEG epoch $\mathbf{x}_2$ randomly;\\
Find the orthogonal direction by Gram-Schmidt process:
$\mathbf{x}_2=q \cdot (\mathbf{x}_2 - \langle \mathbf{x}_1, \mathbf{x}_2 \rangle / \langle \mathbf{x}_1, \mathbf{x}_2 \rangle \times \mathbf{x}_1)$;\\
$\{\mathbf{x}^+,\mathbf{x}^-\}=BinarySearch(\{\mathbf{x}_b^+,\mathbf{x}_b^-\},f',m)$;\\
$\mathbf{x}_s=\mathbf{x}_2+(\mathbf{x}^{+} + \mathbf{x}^-)/2$.\\
~\\
\KwRet $\mathbf{x}_s$
\caption{
    Mid-perpendicular synthesis.
    $\mathbf{x}_s=MidPerp(\{\mathbf{x}_b^+,\mathbf{x}_b^-\},f',k,q)$
}\label{alg:Perp}
\end{algorithm}

With this query-syntheses-based active learning strategy, the entire substitute model training process for binary classification is shown in Algorithm~\ref{alg:SubTraining}.

\begin{algorithm}
\DontPrintSemicolon
\KwIn{$f$, the target model; $S_0$, a set of unlabeled EEG epochs; $N_{\max}$, maximum number of training epochs; $n_{\max}$, maximum number of synthesized EEG epochs in one iteration; $m$, maximum number of binary search iterations; $q$, magnitude of the orthogonal vector.}
\KwOut{$f'$, a trained substitute model\\
~\\}
Label $S_0$ by querying $f$ to obtain an initial training set $D$;\\
Initialize $f'$ and pre-train $f'$ on $D$; \\
$\Delta D=\varnothing$;\\
\For{$N=1$ to $N_{\max}$}
{
    $\Delta S=\varnothing$;\\
    \For{$n=1$ to $n_{\max}$}
    {
        Select an opposite pair $\mathbf{x}_0^+$ and $\mathbf{x}_0^-$ randomly from $D$;\\
        $\{\mathbf{x}_b^+,\mathbf{x}_b^-\}=BinarySearch(\{\mathbf{x}_0^+,\mathbf{x}_0^-\},f',m)$;\\
        $x_s=MidPerp(\{\mathbf{x}_b^+,\mathbf{x}_b^-\},f',m,q)$;\\
        $\Delta S \leftarrow \Delta S \bigcup \{\mathbf{x}_s\}$;\\
    }
    $\Delta D=\{(\mathbf{x}_i,f(\mathbf{x}_i))\}_{\mathbf{x}_i \in \Delta S}$;\\
    $D \leftarrow D \bigcup \Delta D$;\\
    Train $f'$ on $D$;
}
~ \\
\KwRet $f'$
\caption{Query-synthesis-based substitute model training strategy.}\label{alg:SubTraining}
\end{algorithm}

We then extend query-synthesis-based augmentation method to multi-class classification, by the simple one-vs-one approach, which decomposes a multi-class task into multiple binary classification tasks. More specifically, if there are $k_1$ classes, then we can solve $k_2=k_1(k_1-1)/2$ binary classifications instead, and use our active learning strategy to synthesize $n_{\max}/k_2$ EEG epochs (where $n_{\max}$ is the maximum number of synthesized EEG epochs in one iteration) for each binary task.

\subsection{Adversarial Example Crafting for the Target Model}

After training the substitute model, we can generate adversarial examples from it for the target model. Goodfellow \emph{et al.}~\cite{FGSM} proposed to construct adversarial perturbations in the following way:
\begin{align}
    \delta = \varepsilon \cdot \mbox{sign}(\nabla_{\mathbf{x}_i} J(\boldsymbol{\theta},\mathbf{x}_i, y_i)), \label{eq:FGSM}
\end{align}
where $\boldsymbol{\theta}$ are the parameters of the target model $f$, and $J$ the loss function. The main idea is to find an optimal max-norm perturbation $\delta$ constrained by $\varepsilon$ to maximize $J$. The requirements in (\ref{eq:bound}) holds if $\varepsilon \leq \epsilon$ and $l_\infty$-norm is used as the distance metric.

Let $\varepsilon=\epsilon$ so that we can perturb $\mathbf{x}_i$ at the maximum extent. Then, the adversarial example $\mathbf{x}_i^*$ can be re-expressed as:
\begin{align}
    \mathbf{x}_i^* = \mathbf{x}_i + \epsilon \cdot \mbox{sign}(\nabla_{\mathbf{x}_i} J(\boldsymbol{\theta},\mathbf{x}_i, y_i)). \label{eq:GeneratorFGSM}
\end{align}

UFGSM~\cite{zhang2019vulnerability} is an unsupervised extension of FGSM, which replaces the label $y_i$ in (\ref{eq:GeneratorFGSM}) by $y_i'=f(\mathbf{x}_i)$. Then, $\mathbf{x}_i^*$ in UFGSM can be written as:
\begin{align}
    \mathbf{x}_i^* = \mathbf{x}_i + \epsilon \cdot \mbox{sign}(\nabla_{\mathbf{x}_i}J(\boldsymbol{\theta},\mathbf{x}_i, y'_i)).
    \label{eq:GeneratorUFGSM}
\end{align}

UFGSM was used in this paper to construct the adversarial examples.

\section{Experiments}\label{sect:Experiment}

This section presents the experimental results to demonstrate the effectiveness of the query-synthesis-based active learning strategy in transferability-based black-box attacks. We evaluate the vulnerability of three CNN classifiers in EEG-based BCIs.

\subsection{Experimental Setup}

The three BCI datasets, P300 evoked potentials (P300) \cite{EPFLP300}, feedback error-related negativity (ERN)\footnote{https://www.kaggle.com/c/inria-bci-challenge}~\cite{ERN}, and motor imagery (MI)\footnote{http://www.bbci.de/competition/iv/}~\cite{MI4C}, used in our recent study \cite{zhang2019vulnerability} were used again in this study. The data pre-processing steps were also identical.
Three CNN classifiers, EEGNet \cite{EEGNet}, DeepCNN \cite{MNE}, and ShallowCNN \cite{MNE} were used as the target models in our experiments, as in \cite{zhang2019vulnerability}. Adam optimizer~\cite{Adam}, cross-entropy loss function, and early stopping were used in training. Moreover, we applied weights to different classes to address the class imbalance problem in P300 and ERN.

Raw classification accuracy (RCA) and balanced classification accuracy (BCA) were used to evaluate the attack performance, where RCA is the unweighted overall classification accuracy on the test set, and BCA is the average of the individual RCAs of different classes.

In order to simulate the black-box scenario, we partitioned the three datasets into two groups, as described in~\cite{zhang2019vulnerability}: the larger group A of 7/14/7 subjects in P300/ERN/MI was used to simulate the unknown data from the target model, where 80\% epochs were for training the target model and the remaining 20\% for testing. The other smaller group B of 1/2/2 subjects was used to initialize set $S_0$ for training the substitute model. In this way, the 8/16/9 subjects in three datasets can be partitioned in 8/120/36 different ways. We performed both one-division and multi-division black-box attacks. In the one-division experiment, the attacks were repeated 5 times to reduce randomness. In the multi-division experiment, we repeated each division 5 times on P300 and only one time on ERN and MI; so, in total we had 40/120/36 evaluations on the three datasets, respectively.

\subsection{Baseline}

We first evaluated the baseline performance of the three CNN target models on the unperturbed EEG data, as shown in the first part of Table~\ref{tab:results}. Because the MI dataset had 4 classes whereas P300 and ERN had only two, its RCAs and BCAs were much lower.

\renewcommand{\arraystretch}{1.2}
\begin{table*}[htpb] \centering \setlength{\tabcolsep}{2.6mm}
\caption{Average RCAs/BCAs (\%) of Different Target Classifiers on the Three Datasets before and after Black-Box Attacks. }   \label{tab:results}
\begin{tabular}{c|c|c|c|c|c|ccc}
\toprule
\multirow{2}{*}{Experiment}   &\multirow{2}{*}{Dataset}    &\multirow{2}{*}{Target Model $f$}  &\multicolumn{2}{c|}{Baselines}  &\multirow{2}{*}{Method}  &\multicolumn{3}{c}{Substitute Model $f'$}   \\ \cline{4-5}\cline{7-9}
&  &   &Original  &Noisy    &    &EEGNet         &DeepCNN        &ShallowCNN\\
\midrule
\multirow{18}{*}{One-Division} &\multirow{6}{*}{P300} &\multirow{2}{*}{EEGNet}  &\multirow{2}{*}{73.81/72.71}  &\multirow{2}{*}{73.78/72.46}  &Ours &\textbf{54.90}/\textbf{54.63}  &\textbf{41.39}/\textbf{50.80} &72.71/\textbf{71.79} \\ & & & & &Jacobian-based  &62.36/61.64  &54.58/55.55  &\textbf{72.07}/72.02 \\ \cline{3-9}
                               &                      &\multirow{2}{*}{DeepCNN} &\multirow{2}{*}{77.77/74.79}  &\multirow{2}{*}{78.08/74.78}  &Ours &69.77/70.52 &\textbf{47.41}/\textbf{59.00}  &\textbf{74.86}/\textbf{73.38} \\ & & & & &Jacobian-based  &\textbf{66.40}/\textbf{66.82}  &53.11/60.13  &76.99/74.20 \\ \cline{3-9}
                               &                      &\multirow{2}{*}{ShallowCNN} &\multirow{2}{*}{72.56/72.87}  &\multirow{2}{*}{72.64/72.87} &Ours &\textbf{63.43}/\textbf{65.17}  &\textbf{60.84}/\textbf{62.18}  &\textbf{66.54}/\textbf{67.91} \\  & & & & &Jacobian-based &67.52/67.74  &74.20/66.15  &73.24/71.73 \\ \cline{2-9}
                               &\multirow{6}{*}{ERN}  &\multirow{2}{*}{EEGNet}  &\multirow{2}{*}{76.26/74.19}  &\multirow{2}{*}{75.47/73.34}  &Ours &\textbf{36.55}/\textbf{38.23}  &\textbf{36.55}/\textbf{41.10}  &\textbf{73.56}/\textbf{71.76} \\  & & & & &Jacobian-based &52.07/51.65  &47.06/50.94  &75.82/72.37 \\ \cline{3-9}
                               &                      &\multirow{2}{*}{DeepCNN} &\multirow{2}{*}{77.21/76.85}  &\multirow{2}{*}{75.05/76.38}  &Ours &\textbf{46.22}/\textbf{46.57}  &\textbf{52.07}/\textbf{61.59}  &\textbf{75.14}/\textbf{73.99} \\  & & & & &Jacobian-based &51.65/52.79  &55.36/64.33  &75.67/75.97 \\ \cline{3-9}
                               &                      &\multirow{2}{*}{ShallowCNN} &\multirow{2}{*}{71.85/71.17}  &\multirow{2}{*}{71.51/70.95} &Ours &\textbf{70.69}/\textbf{71.44}  &\textbf{69.75}/\textbf{70.66}  &\textbf{44.82}/\textbf{49.81} \\  & & & & &Jacobian-based &72.16/71.50  &70.73/71.22  &52.10/53.57 \\ \cline{2-9}
                               &\multirow{6}{*}{MI}   &\multirow{2}{*}{EEGNet} &\multirow{2}{*}{54.93/54.62}  &\multirow{2}{*}{52.93/52.78}   &Ours &\textbf{31.77}/\textbf{31.70}  &35.96/36.07  &\textbf{40.27}/\textbf{40.28} \\  & & & & &Jacobian-based &37.07/36.87  &\textbf{35.26}/\textbf{35.24}  &41.01/41.02 \\ \cline{3-9}
                               &                      &\multirow{2}{*}{DeepCNN} &\multirow{2}{*}{49.38/49.18}  &\multirow{2}{*}{49.73/49.56}   &Ours &\textbf{41.91}/\textbf{41.78}  &\textbf{39.24}/\textbf{39.08}  &\textbf{38.14}/\textbf{38.05} \\  & & & & &Jacobian-based &45.36/45.22  &40.56/40.48  &41.09/41.06 \\ \cline{3-9}
                               &                      &\multirow{2}{*}{ShallowCNN} &\multirow{2}{*}{60.96/60.81}  &\multirow{2}{*}{61.04/60.93}  &Ours &\textbf{51.93}/\textbf{52.06}  &51.93/52.01  &\textbf{43.60}/\textbf{43.68} \\  & & & & &Jacobian-based &55.05/55.18  &\textbf{48.28}/\textbf{48.44}  &48.97/49.00 \\
\midrule
\multirow{18}{*}{Multi-Division} &\multirow{6}{*}{P300} &\multirow{2}{*}{EEGNet}  &\multirow{2}{*}{73.59/71.95}  &\multirow{2}{*}{73.52/71.83}  &Ours &\textbf{41.40}/\textbf{42.44}  &\textbf{32.66}/\textbf{39.75}  &\textbf{64.95}/\textbf{64.91} \\ & & & & &Jacobian-based &47.02/48.49  &39.79/41.99  &65.16/64.16 \\ \cline{3-9}
                               &                      &\multirow{2}{*}{DeepCNN} &\multirow{2}{*}{75.99/74.10}  &\multirow{2}{*}{76.20/73.94}  &Ours &\textbf{53.29}/\textbf{55.10}  &\textbf{36.57}/\textbf{44.55}  &\textbf{68.78}/\textbf{66.88} \\  & & & & &Jacobian-based &59.18/60.01  &44.54/49.66  &70.05/66.98 \\ \cline{3-9}
                               &                      &\multirow{2}{*}{ShallowCNN} &\multirow{2}{*}{72.23/71.90}  &\multirow{2}{*}{72.24/71.85} &Ours &\textbf{59.66}/\textbf{60.17}  &51.76/\textbf{55.15}  &53.80/\textbf{49.60} \\  & & & & &Jacobian-based &59.75/61.73  &\textbf{51.40}/57.91  &\textbf{52.38}/52.54 \\ \cline{2-9}
                               &\multirow{6}{*}{ERN}  &\multirow{2}{*}{EEGNet}  &\multirow{2}{*}{73.89/72.94}  &\multirow{2}{*}{73.23/72.72}  &Ours &\textbf{45.71}/\textbf{47.29}  &\textbf{46.78}/\textbf{48.94}  &71.27/71.14 \\  & & & & &Jacobian-based &54.42/54.47  &52.54/54.53  &\textbf{71.22}/\textbf{70.97} \\ \cline{3-9}
                               &                      &\multirow{2}{*}{DeepCNN} &\multirow{2}{*}{74.24/72.69}  &\multirow{2}{*}{73.86/72.38}  &Ours &\textbf{56.78}/\textbf{55.07}  &\textbf{53.85}/\textbf{53.33}  &\textbf{72.44}/\textbf{70.89} \\  & & & & &Jacobian-based &59.64/58.32  &57.77/57.54  &72.73/71.43 \\ \cline{3-9}
                               &                      &\multirow{2}{*}{ShallowCNN} &\multirow{2}{*}{71.86/71.45}  &\multirow{2}{*}{71.77/71.21} &Ours &\textbf{70.15}/70.46  &\textbf{68.49}/\textbf{69.28}  &\textbf{58.28}/\textbf{59.42} \\  & & & & &Jacobian-based &70.44/\textbf{70.30}  &70.31/70.18  &63.51/63.80 \\ \cline{2-9}
                               &\multirow{6}{*}{MI}   &\multirow{2}{*}{EEGNet} &\multirow{2}{*}{60.85/60.71}  &\multirow{2}{*}{59.48/59.39}   &Ours &\textbf{35.10}/\textbf{35.27}  &\textbf{46.47}/\textbf{46.45}  &\textbf{43.60}/\textbf{43.67} \\  & & & & &Jacobian-based &39.13/39.13  &46.58/46.61  &53.19/53.17 \\ \cline{3-9}
                               &                      &\multirow{2}{*}{DeepCNN} &\multirow{2}{*}{55.83/55.57}  &\multirow{2}{*}{55.65/55.39}   &Ours &\textbf{47.06}/\textbf{46.90}  &\textbf{45.32}/\textbf{45.31}  &\textbf{42.54}/\textbf{42.46} \\  & & & & &Jacobian-based &48.68/48.59  &48.45/48.38  &48.72/48.57 \\ \cline{3-9}
                               &                      &\multirow{2}{*}{ShallowCNN} &\multirow{2}{*}{64.71/64.62}  &\multirow{2}{*}{64.18/64.09}  &Ours &\textbf{57.26}/\textbf{57.32}  &\textbf{57.94}/\textbf{57.98}  &\textbf{46.74}/\textbf{46.84} \\  & & & & &Jacobian-based &59.44/59.49  &59.04/59.07  &54.56/54.56 \\
\bottomrule
\end{tabular}
\end{table*}

We then constructed a random perturbation $\delta'$:
\begin{align}
    \delta' = \epsilon \cdot \mbox{sign}\left(\mathcal{N}(0,1)\right), \label{eq:RandomNoise}
\end{align}
which has the same maximum amplitude $\epsilon$ as the adversarial perturbations, to verify the necessity of deliberately constructing the adversarial examples. The results are shown in the second part of Table~\ref{tab:results}. It is obvious that the target models were robust to random noise, i.e., random noise cannot effectively perform adversarial attacks.

\subsection{Attack Performance Comparison}

We next compared our query-synthesis-based approach with the Jacobian-based method~\cite{zhang2019vulnerability} in black-box attacks. $\epsilon=0.1/0.1/0.05$ on P300/ERN/MI were used to construct the adversarial examples after the substitute models were trained.

In one-division experiments, we randomly downsampled the data for each class according to the labels that the target model predicted at the first time to balance the classes in the initial dataset. We set the $downsampled$ $number$ in each class as 200 in P300 and ERN, and 100 in MI, to ensure that the size of the initial substitute model training set $S_0$ was 400. The initial substitute model was then trained on $S_0$ using both the Jacobian-based method and ours. $\lambda=0.5$ and $N=1$ were used in the Jacobian-based method, where $N$ corresponds to $N_{\max}$ in our method. $n_{\max}=200$ and $m=10$ were used in our method, so $N_{\max}=2$ should be used to get the same number of queries as in the Jacobian-based method. $q=1.0/0.8/0.8$ on P300/ERN/MI were used in the experiments. They were determined such that the generated EEG epochs had approximately the same magnitudes as those in the initial training set $S_0$.

In multi-division experiments, we did not limit $downsampled$ $number$ to avoid wasting data, but always kept the same number of queries in the Jacobian-based method and ours. so $n_{\max}=downsampling~number$ on P300 and ERN, $n_{\max}=2*downsampling~number$ on MI in our method, the other hyper-parameters were the same as those in one-division experiments. The attack results are shown in Table~\ref{tab:results}. We can observe that:

\begin{enumerate}
\item Generally, the RCAs and BCAs of the target models after attacks were lower than the corresponding baselines, indicating the effectiveness of the transferability-based black-box attack methods. However, shallowCNN did not perform well in attacking EEGNet and deepCNN on P300 and ERN.
\item The RCAs and BCAs after query-synthesis-based black-box attacks were generally lower than the corresponding accuracy after Jacobian-based black-box attacks. For example, in one-division experiments in which only 400 queries were used for both methods, when the target model and the substitute model were EEGNet and DeepCNN respectively on P300, our method achieved 13\% improvement over the Jacobian-based method.
\item Our proposed method was often much better than the Jacobian-based method when the substitute model and the target model had the same structure, which may due to the strong intra-transferability of the adversarial examples.
\end{enumerate}

\begin{figure*}[htbp]   \centering
\includegraphics[width=\linewidth,clip]{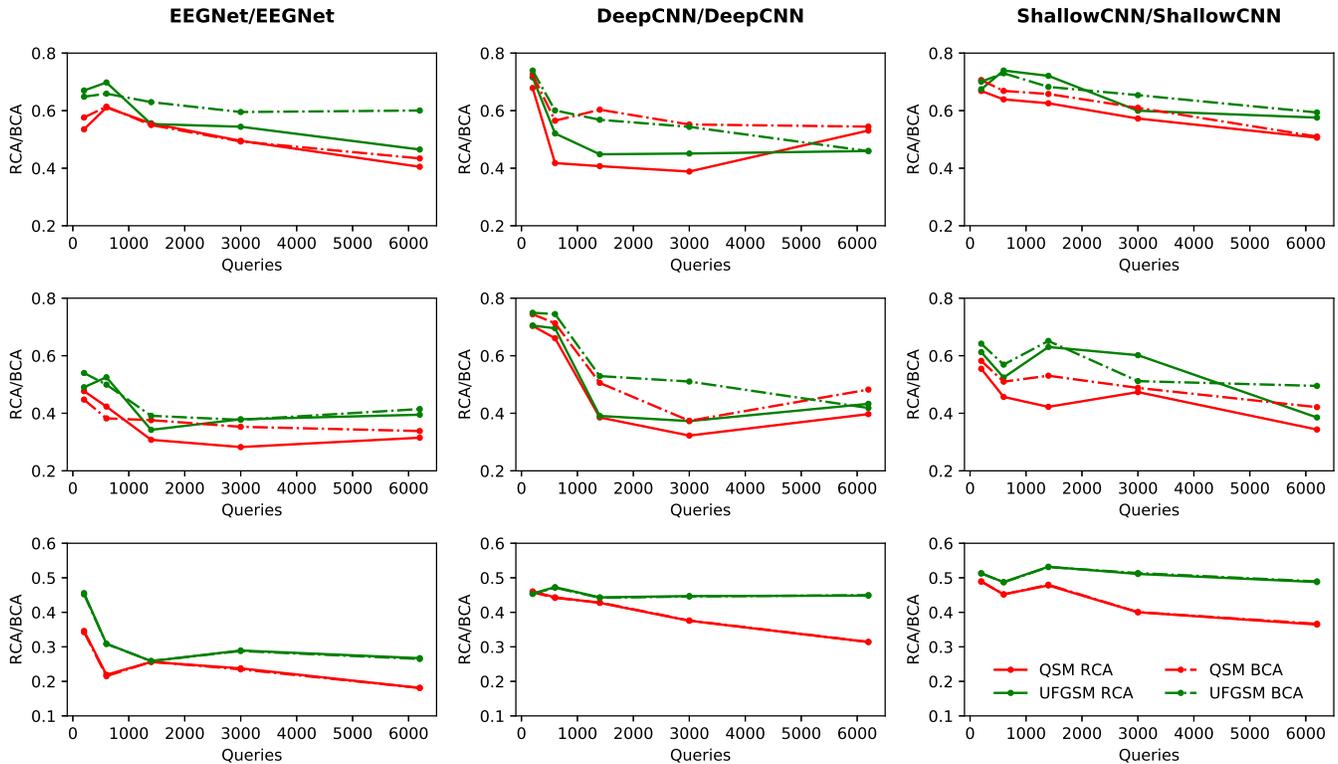}
\caption{RCAs and BCAs of the two black-box attack methods with different number of queries on P300 (top row), ERN (middle row) and MI (bottom row).} \label{fig:difQuiries}
\end{figure*}

We compared the attack performance with different number of queries when the substitute model and the target model had the same architecture, i.e., both were EEGNet, DeepCNN or ShllowCNN. 200 EEG epochs were used for initializing the substitute model, and the number of queries varied from 200 to 6,200. The corresponding RCAs/BCAs are shown in Fig.~\ref{fig:difQuiries}. Generally, both RCAs and BCAs decreased as the number of queries increased, which is intuitive. With the same number of queries, our method achieved lower RCAs and BCAs than the Jacobian-based method, i.e., it gave higher attack success rates. In other words, it can achieve a desired attack success rate with much fewer queries. For example, for EEGNet on P300, our method only needed 3,000 queries to achieve a comparable attack performance with the Jacobian-based method with 6,200 queries.

\subsection{Characteristics of the Adversarial Examples}

In this section, we analyzed the characteristics of the generated adversarial examples in both time domain and spectral domain.

\begin{figure}[htbp]   \centering
\includegraphics[width=\linewidth,clip]{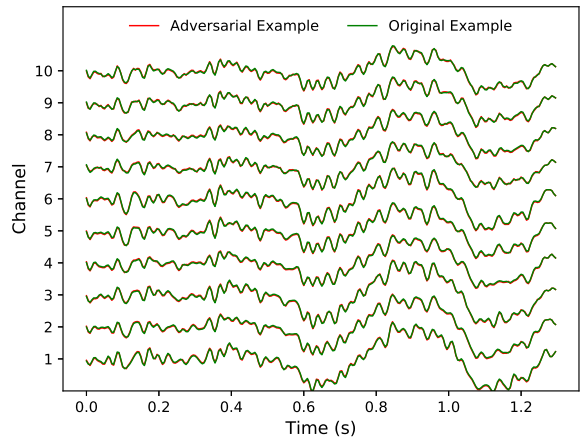}
\caption{Example of an original EEG epoch and its adversarial epoch, generated by query-synthesis-based black-box attack on the ERN dataset. The first 10 channels are shown. $\epsilon=0.1$.} \label{fig:TimeDomain}
\end{figure}

Consider one-division query-synthesis-based black-box attacks on ERN after 400 queries, when both the target model and the substitute model are DeepCNN. An example of the original EEG epoch (we only show the first 10 channels) and its corresponding adversarial epoch is shown in Fig.~\ref{fig:TimeDomain}. They were almost completely overlapping in the time domain, which means the adversarial example is very difficult to be detected by human or a computer. We had similar observations on other datasets and from other classifiers.

\begin{figure*}[htbp]   \centering
\includegraphics[width=\linewidth,clip]{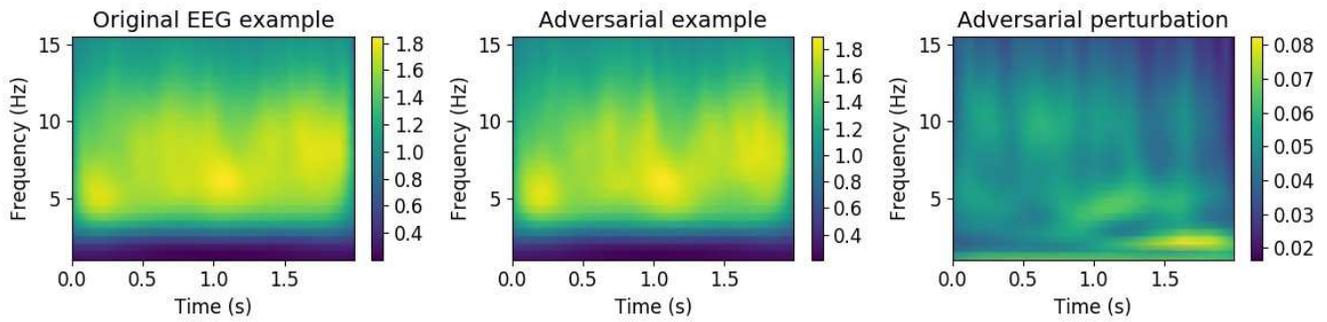}
\caption{Mean spectrogram of the original EEG epochs, of all successful adversarial examples, and of the corresponding perturbations, in one-division query-synthesis-based black-box attacks on the MI dataset. Channel $F_z$ was used.} \label{fig:SpecDomain}
\end{figure*}

Next, spectrogram analysis was used to further explore the characteristics of the adversarial examples. Fig.~\ref{fig:SpecDomain} shows the mean spectrogram of all original EEG epochs, of all adversarial examples whose predicted labels did not match the ground-truth labels of the original EEG epochs (successful attacks), and of the corresponding successful perturbations ($\mathbf{x}^* - \mathbf{x}$), by using wavelet decomposition. The adversarial examples were crafted by query-synthesis-based black-box attacks on MI. Both the target and the substitute models were DeepCNN. There was no significant difference in the spectrograms of the original EEG epochs and the adversarial examples. The amplitudes of the mean spectrogram of the adversarial perturbations were much smaller than those of the original and adversarial examples, suggesting that our generated adversarial examples are difficult to be detected by spectrogram analysis. We had similar observations on other datasets and from other classifiers.

\section{Conclusion}\label{sect:Conclusion}

In this paper, we have proposed a query-synthesis-based active learning strategy for transferability-based black-box attacks. It improves the query efficiency by actively synthesizing EEG epochs scattering around the decision boundary of the target model, and thus the trained substitute model can better approximate the target model. We applied our method to attacking three state-of-the-art deep learning classifiers in EEG-based BCIs. Experiments demonstrated its effectiveness and efficiency. With the same number of queries, it can achieve better attack performance than the traditional Jacobian-based method; or, in other words, our approaches needs a smaller number of queries to achieve a desired attack success rate.

%\bibliographystyle{IEEEtran}\bibliography{xjiangbib}
% Generated by IEEEtran.bst, version: 1.14 (2015/08/26)

\end{document}